\documentclass[pdflatex,sn-mathphys]{sn-jnl}

\jyear{2021}%

\theoremstyle{thmstyleone}%
\theoremstyle{thmstyletwo}%

\theoremstyle{thmstylethree}%

\usepackage{tabularx}

\begin{document}

\title[Lemmatization and Morphological Tagging Method in Turkish]{Context Aware Lemmatization and Morphological Tagging Method in Turkish}


\author*[1]{\fnm{Çağrı} \sur{Sayallar}}\email{190201077@kocaeli.edu.tr}

\affil*[1]{\orgdiv{Computer Engineering}, \orgname{Kocaeli University}, \orgaddress{\street{Umuttepe}, \city{Kocaeli}, \postcode{41001}, \state{İzmit}, \country{Turkey}}}


\abstract{The smallest part of a word that defines the word is called a word root. Word roots are used to increase success in many applications since they simplify the word. In this study, the lemmatization model, which is a word root finding method, and the morphological tagging model, which predicts the grammatical knowledge of the word, are presented. The presented model was developed for Turkish, and both models make predictions by taking the meaning of the word into account. In the literature, there is no lemmatization study that is sensitive to word meaning in Turkish. For this reason, the present study shares the model and the results obtained from the model on Turkish lemmatization for the first time in the literature. In the present study, in the lemmatization and morphological tagging models, bidirectional LSTM is used for the spelling of words, and the Turkish BERT model is used for the meaning of words. The models are trained using the IMST and PUD datasets from Universal Dependencies. The results from the training of the models were compared with the results from the SIGMORPHON 2019 competition. The results of the comparisons revealed that our models were superior.}




\maketitle

\section{Introduction}\label{sec1}

A root is the lowest, most significant component of a word. Due to the agglutinative nature of the Turkish language, the same root word can appear in many words. For this reason, in applications such as sentence analysis, machine translation, and small datasets, the root of the word is utilized instead of the actual word. According to the meaning of the word, the two word root-finding methods are categorized as stemming and lemmatization. The stemming method is an algorithm that reduces the word to smaller parts than predetermined prefixes. The reason why it is called the "smaller part" is that stemming algorithms cannot always obtain the root of the word. In other words, it guesses a completely meaningless word fragment. In the stemming method, the content of the sentence is not considered when the root of the word is found. Words are evaluated independently of the sentence and word roots are sought. For this reason, it cannot accurately predict ambiguous words that have the same spelling but whose root words will change according to their meaning in the sentence. For example, while the root of the word "oku" in Turkish can be the verb "oku(-mak)", it can also be derived from the noun "ok". Without the context of the statement, stemming can only speculate against such confusing words. The lemmatization method has been produced as a solution to the problem of the stemming method, which is unaware of the sentence content and the meaning of the word. The proposed lemmatization method finds the root word while taking into account the meaning of each word in the phrase. In the lemmatization approach, a word's root can vary from sentence to sentence, but in the stemming method, words written in the same way always produce the same root.  

The purpose of this study is to develop a word-meaning-sensitive lemmatization and morphological tagging method for Turkish individuals. Depending on whether it relies on the results of the morphological tagging model, the developed lemmatization model consists of two alternative models. Words are morphologically classified into categories such as noun, verb, adjective, pronoun, number, punctuation mark, and gender. A word can have more than one morphological tag at the same time. The morphological tag of a word can change according to the meaning of the word, as in the lemmatization method. For example, the word "oku" can be labeled a noun or a verb depending on its meaning in the sentence. Here, correct guessing of the morphological tag of the word can also help the lemmatization process, as seen from the example given. For this reason, studies use the output of one of these two tasks as the input of the other, but when the mentioned studies are examined, sending the output obtained from the morphological labeling model to the lemmatization model does not always increase success \cite{muller_joint_2015,kondratyuk_lemmatag_2018,malaviya_simple_2020}. In our study, two distinct lemmatization models were used, and the results were given in accordance with the input of the morphological tag for the lemmatization model, since it is difficult to find datasets with the morphological tag of the word. In lemmatization and morphological tagging models, the meaning of the word must be known. In both models, the word vector obtained from the Turkish Bidirectional Encoder Representations from Transformers (BERT) \cite{devlin_bert_2019} model was used to describe the meaning of the word. Bidirectional long-short-term memory (LSTM) \cite{lstm} is used to represent the spelling of a word. One-way LSTM is used for the generation of root words or morphological tags. Similar models are used for both methods.

The results of the SIGMORPHON 2019 \cite{mccarthy_sigmorphon_2019} competition are used to compare the results of the developed model. Our work focuses on the second task in the SIGMORPHON 2019 competition. The mentioned task consists of the generation of morphological tags and word stems via data sets from the Universal Dependencies Database \cite{nivre_universal_2020}. In the competition, the achievements of the participants are compared in 100 Universal Dependencies datasets consisting of 66 languages, including Turkish. As a result of the competition, the highest achievement is shared in each dataset and each comparison metric. Among the shared results are the achievements of the Universal Dependencies IMST and PUD datasets in Turkish. The success of this study is compared to the metrics determined by the SIGMORPHON competition in the IMST and PUD datasets mentioned. Our work achieves the highest success on two datasets and almost all metrics. 

Lemmatization and stemming studies in Turkish are first referenced in the literature in the following Literature Review section. The most well-known studies on lemmatization and morphological tagging are described below. The morphological labeling and lemmatization architecture created for Turkish is mentioned in the Methods section. The following section presents the "Experiments and Results" section and presents the dataset and test results. The study is outlined, and the results are discussed in the Conclusions and Comments section. 

\section{Literature Review}\label{sec2}

Many algorithms have been proposed for stemming methods that break words into smaller parts. The most well-known of these is the Snowball \cite{Snowball_Stemming} stemming method, which is based on the Porter \cite{Porter_Algorithm_1980} algorithm. Snowball divides the word into smaller parts by removing the suffixes at the end. Although the Snowball algorithm often yields better results than other stemming algorithms, it still cannot predict the root word correctly. In Turkish, for example, the root of the word "bakan" can be either the verb "bak(-mak)" or the noun "bakan." When the mentioned word is sent to the Turkish Snowball algorithm, its root is estimated as "baka", and this word is a meaningless word that does not exist in Turkish. In addition to the stemming algorithms, studies have been carried out only for the Turkish language. In one of these studies \cite{silahtaroglu_lemmatizer_2020}, a finite-state machine (FSM) was developed to find the root of Turkish words. The developed method involves removing the suffixes of the word to identify the root of the word. They used 130 different predetermined suffixes from the Turkish language when deciding how to affix the noun. In this study, root estimation is performed according to the spelling of the word, not the meaning of the word, so their work is similar to that of the stemming algorithms. In the study \cite{can_lstm_2019}, a deep learning model was used for root prediction. By running the word letter-by-letter through the bidirectional LSTM during the model's encoder stage, they are able to create the vector used to represent the word. The word's root is then produced as letters as they proceed through the forward LSTM stage of the decoder. They presented two different models and their results according to the use of attention in the decoder stage. Like his study \cite{silahtaroglu_lemmatizer_2020}, only the spelling of the word is taken into account in the training of the models. This situation makes the ambiguous words mentioned difficult to guess correctly. In addition to these studies, the Zemberek study \cite{af_zemberek_nodate} described the natural language processing framework developed for Turkish. Zemberek's study includes morphological analyses of the word as well as the process of determining the root of the word. While the root of the word is being found, first, candidates that can be the root of the word are being produced. Among the candidates obtained later, the candidate with the highest value according to the frequency of use of the word is selected as the root. Although this method predicts the root of ambiguous words, it ensures that it always predicts the same root regardless of the meaning of the word. This problem will be solved in future studies. Unlike the aforementioned Turkish studies \cite{af_zemberek_nodate,can_lstm_2019,silahtaroglu_lemmatizer_2020}, the meaning of the word in the sentence was also taken into account in our study. As a result, an attempt has been made to solve the root prediction problem of ambiguous words.

There are studies supported by multiple languages, including Turkish, in addition to those that are specifically focused on Turkish. In the Universal Dependencies dataset, Lematus \cite{bergmanis_context_2018} illustrates the successes of the model they created in 20 different languages. In this study, the LSTM network is used to represent the meaning of the word in the sentence by passing the characters before and after the word whose root must be determined. In this study, which compares its success with this study and shares its success in Turkish \cite{malaviya_simple_2020}, bidirectional LSTM is used for the meaning of the word in the sentence. All words in a sentence are first passed through the LSTM on a character basis. Here, the LSTM hidden state represents the vector of the word. Each word vector in the sentence is then passed through the word-based bidirectional LSTM. Thus, the meaning of each word in the sentence is represented as a vector. In the two studies mentioned \cite{bergmanis_context_2018,malaviya_simple_2020} when the root of the word was predicted, other words in the sentence were also used, similar to our study. 

Recently, there has been competition for morphological tagging and lemmatization, similar to those in multilingual studies. As stated in the introduction and task 2 of the SIGMORPHON 2019 competition, the model that performs the best in 66 distinct languages and 100 different datasets is chosen. The aim of this competition is to increase the success by transferring morphological information from languages with a large number of resources to languages with few resources. The Universal Dependencies dataset is used in the competition. The lemmatization accuracy (i), the Levenshtein distance (ii), the morphological labeling accuracy (iii), and the morphological labeling f-1 score metrics were measured with the data set (iv). The achievements of the participants are compared with those of the studies with and without artificial intelligence, which were determined both among themselves and by the contest owners. As a result of the competition, the best performance among the participants was shared for each language and each separate metric mentioned. Among the shared results are the best results in the Universal Dependencies IMST and PUD datasets for the four different metrics mentioned. To measure the success of this study, it is compared with the mentioned results. Since the models in the competition in which the comparison was made were developed for multiple languages and the method described in this study focuses only on the Turkish language, the comparison may be considered unfair, but there is no word-meaning sensitive lemmatization study in the Turkish language. For this reason, the comparison is made with the best multilingual, context-aware study that includes Turkish; among these studies \cite{bergmanis_context_2018,mccarthy_sigmorphon_2019,malaviya_simple_2020} the SIGMORPHON 2019 competition had the best results in Turkish.

In this study, unlike the studies mentioned above, morphological tagging and lemmatization methods are presented, which focus on the Turkish language and consider the meaning of the word in the sentence. In our study, the BERT model was used to determine the meaning of words in the sentence \cite{bergmanis_context_2018,kondratyuk_cross-lingual_2019,kondratyuk_75_2019,straka_udpipe_2019}. In some cases, the output of the morphological tagging model increases the success of the lemmatization model. As a result, the estimated morphological tags in our study are used as inputs in the lemmatization model, as in previous studies \cite{muller_joint_2015,kondratyuk_lemmatag_2018,malaviya_simple_2020}, and the estimated morphological tags in our study are used as inputs in the lemmatization model. In addition, the results of the lemmatization model, which is independent of the morphological tagging model, are shared because it is difficult to find the data set containing the morphological tag.

\section{Method}\label{sec3}

Lemmatization and morphological tagging are comparable techniques. Both the word's spelling and its meaning must be accurately reflected for the prediction to be accurate. Since the approaches for solving morphological tagging and lemmatization problems are similar, similar models are used for both problems in this study. The only difference between the morphological tagging and lemmatization models is that the lemmatization model takes morphological tags as input. Similar models are used for both problems in this work, since the approaches for solving the morphological tagging and lemmatization problems are similar. The only difference between morphological tagging and lemmatization is that the lemmatization model can use morphological tags as input. The model, which is independent of the lemmatization and morphological tagging model, is called a separate model. A diagram of the sequenced and separate models is shown in Figure 1. The morphological tagging model and the lemmatization model are trained independently in the separate model. For the sequenced model, the morphological labeling model was trained first, and the model with the highest success rate was obtained. The lemmatization model is then trained using the morphological tag predictions as input that were made by this model. The morphological labeling model in this architecture, known as the sequenced model in our study, remains unchanged while the lemmatization model is trained. Therefore, the results of the morphological tagging models given in Table 1 and Table 2 are exactly the same. The sequenced model's only advantage is that, in some cases, it increases success rates in the lemmatization model. The morphological tagging and lemmatization models used here are described as a single model. First, the BERT model used for word meaning is mentioned. The structure used in the morphological tagging and lemmatization models is mentioned in the encoder section. In the decoder stage, the estimated output is mentioned. 


\begin{figure}[h]
\centering
\includegraphics[scale=0.39]{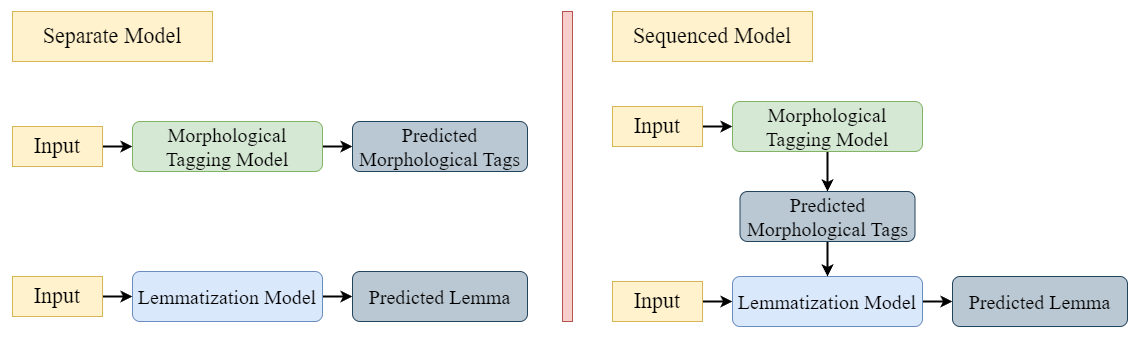}
\caption{Separate Model and Sequenced Model.}
\end{figure}


To address words that are unknown, the created model represents them as characters. Predictions are made one word at a time. The character-level representation of the word to be predicted and the output of the sentence in which the word is found via the BERT model tokenizer must both be provided as inputs to the model. The word's character input corresponds to its spelling, whereas the BERT model's output corresponds to its definition. Turkish DistilBERT \cite{schweter__2022}, which was prepared for the Turkish language and is case sensitive, is used as the BERT model. The BERT model takes the sentence passed through the tokenizer used in its training as input. As output, it returns the vector of each word or part of the word in the sentence. The BERT model divides the word into parts that it knows to cope with the words it does not know. Thus, it can represent a single word with more than one word fragment. The output of the BERT model consists of words and word parts in the sentence. For this output to represent a single word, the word vector or, if the word is fragmented, the sum of the word segment vectors is used.

\subsection{Encoder}\label{subsec1}
The encoder part of the model is responsible for generating the spelling vector and the meaning vector of the word. For word meaning, a single vector is obtained by applying a mask to the outputs obtained from the BERT model, as mentioned before. Character embedding and bidirectional LSTMs are used for the word spelling vector. Word characters are represented by numbers. The one-dimensional vector representing this word is passed through the embedding layer after being input into the model. Thus, each word is now represented by two-dimensional vectors. 


\begin{figure}[h]
\centering
\includegraphics[scale=0.42]{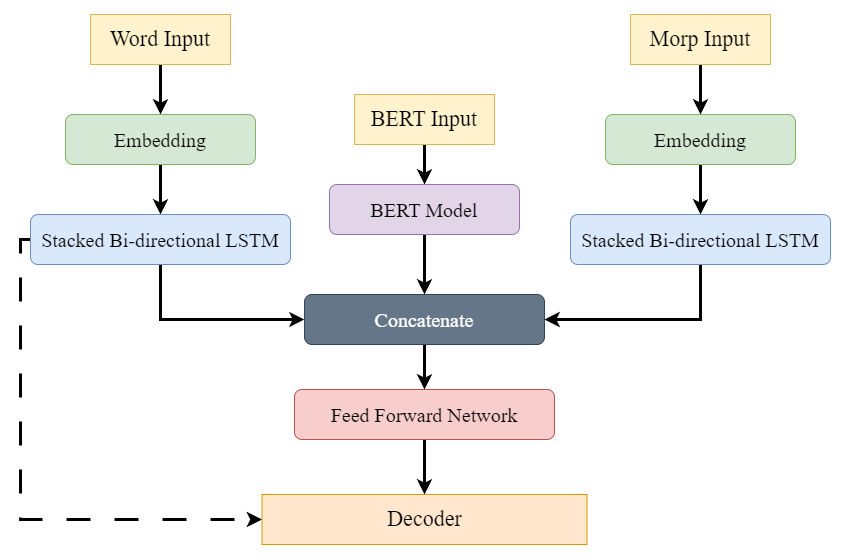}
\caption{Encoder scheme of the model used in the lemmatization and morphological tagging model. The arrow represents the layer output. The dashed arrow represents the hidden state output of the last Bi-LSTM layer.}
\end{figure}


The output of the embedding layer is passed through the bidirectional LSTM layer. This LSTM output is passed through the next LSTM, and the two outputs are added together and sent to the third LSTM. The third LSTM output, which is the last layer, is combined with the previous LSTM output. While the hidden state obtained from the last LSTM is the vector representing the spelling of the word, the LSTM output represents each letter of the word. Unlike the morphological tagging model, the lemmatization model can accept morphological tags as input, as demonstrated in Figure 1. The results of the morphological tag model produce these morphological tags. For incoming morphological tags, embedding and a bidirectional LSTM structure specific to morphological tags are used. This structure is indicated in Figure 2 as the sequenced lemmatization model. The hidden state in the last layer of the LSTM created for the morphological tags represents the morphological tags as vectors. Finally, the word-selling vector, word-meaning vector and morphological tag vector obtained according to the model are passed through dense layers, and a single vector representing all the information is obtained. Later, this vector is used as a hidden state input in the decoder stage. The encoder scheme described is shown in Figure 2. 

\subsection{Decoder}\label{subsec2}

The information from the encoder layer is used to produce outputs in the decoder layer step by step. These outputs are morphological tags in the morphological tagging model and letters in the word root in the lemmatization model. In the decoder section, two forward LSTM layers, an attention layer, dense layers, and a normalization layer, are used to produce these outputs. As the first LSTM input, the expected output is shifted one unit to the right, as in the Teacher Forcing method. The LSTM input for the first step is the special token "<start>". The only vector that contains data from the Word Meaning Vector, Word Spelling Vector, and Morphological Tags Vector in the Sequenced Lemmatization Model, formed at the encoder step, is used as the hidden state input of the first LSTM. This vector contains information about the spelling of the word and the meaning of the word. While the first LSTM output is used as the input for the second LSTM, the hidden state is transferred between the LSTM layers. The two LSTM outputs obtained are passed through the addition and normalization layers. After the new output passes through the Bahdanau Attention \cite{bahdanau_neural_2016} layer with the output created in the encoder section, which is the last LSTM output and contains the information for each letter of the word, the addition and normalization processes are then applied. The addition and normalization process is then applied once more after the newly obtained output has been transmitted through the dense layer. Finally, the model output is produced in a single step by passing it to the classification layer. The output from the classification layer is transformed into a one-hot vector and used as the next step input, with the final LSTM hidden state from the decoder layer acting as the first LSTM hidden state. This cycle is repeated until all the output steps are completed. The decoder scheme is depicted in Figure 3. 


\begin{figure}[h]
\centering
\includegraphics[scale=0.53]{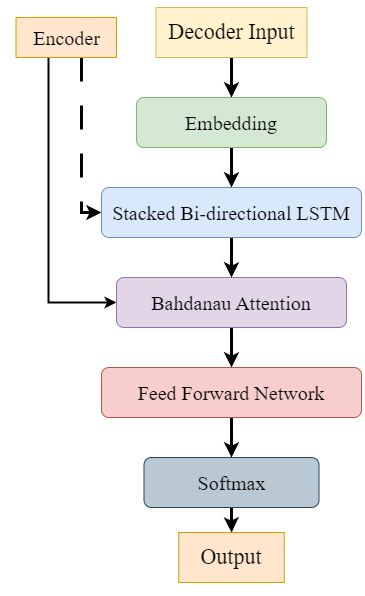}
\caption{Decoder scheme of the model used in the lemmatization and morphological tagging model. The arrow represents the layer output. The dashed arrow represents the hidden state output of the last Bi-LSTM layer.}
\end{figure}


\section{Experiments and Results}\label{sec4}

This section first describes the data set and then describes the training parameters used to train the suggested model. The training results are then evaluated by comparing them with the outcomes of the SIGMORPHON competition. 

\subsection{Dataset}\label{subsec3}

The Universal Dependencies Turkish IMST and PUD datasets are used with our model. The version of the datasets shared by the SIGMORPHON competition is used for a fair comparison of model results. The data set includes each word in the sentence, the root of the word, and its morphological tags. Since the data set is in Conllu format, the Conllu library \cite{stenstrom_conll-u_2022} is used for processing. Because there are different numbers of morphological tag classes in the IMST and PUD datasets, different predetermined morphological tag lists were established for both datasets when tagging the data. The word morphological tags are arranged in order before they are included in the training data because the morphological tags are mixed in both datasets and their order is irrelevant. More success is achieved with sequential labeling of morphological tags than with mixed tagging. 

\subsection{Training Parameters}\label{subsec4}

There are two phases in model training. The BERT model is frozen during the first step of training. The BERT model is also trained with a slower learning rate in the second stage. To greatly reduce training time, the output of the BERT model is created prior to training and sent as input to the model in the first stage of training. The BERT model is also trained during the second stage. 

In the first stage of training, the model is trained for 128 epochs. The weight that gives the best success is recorded. The value 64 is used as the batch size. The Adam function \cite{kingma_adam_2017} was used for the optimization function, and the 1E-3 value was used for the learning rate. Categorical cross-entropy was used as the loss function, whereas Softmax was used for classification. The weight with the highest degree of success from the first phase of training is used to continue training in the second phase. At this point, the BERT model is trained with a batch size of 32 and a learning rate of 1e-5. Training is stopped early if the second stage has been trained for 48 epochs and has successfully completed the required number of epochs in a row under a particular loss value. Once more, the best result is recorded. 

\subsection{Tests}\label{subsec5}

The model output contrasts with the results of the SIGMORPHON 2019 competition. The competition owners made available the datasets that were used in the contest. The data sets were obtained from the Universal Dependencies database; however, the contest owners modified the data sets. For this reason, the Universal Dependencies Turkish IMST and PUD datasets shared in the competition were used to compare the results. These data sets are compared across the categories determined in the SIGMORPHON competition. The code by which the success of our model is measured is the benchmark code shared by the SIGMORPHON competition. The achievements of our model and the best achievements in the IMST dataset in the SIGMORPHON competition are shared in Table 1, and the achievements in the PUD dataset are shared in Table 2. 


\begin{table}[h]
    \centering
    \caption{Results of the IMST dataset.}
    \begin{tabular}{ | m{2cm} | m{19mm} | m{19mm} | m{19mm} | m{19mm} |} 
      \hline
      Model & Lemmatization Accuracy & Levenshtein Distance & Morphological Accuracy & Morphological F-1 Score \\ 
      \hline
      SIGMORPHON & 96.84 & 0.06 & 92.27 & 96.30 \\
      \hline
      \textbf{Separate Model} & 97.70 & 0.03 & \textbf{93.75} & \textbf{98.24} \\
      \hline
      \textbf{Sequenced Model} & \textbf{98.13} & \textbf{0.02} & \textbf{93.75} & \textbf{98.24}  \\
      \hline
    \end{tabular}
    \label{tab:table_1}
\end{table}


\begin{table}[h]
    \centering
    \caption{Results of the PUD dataset.}
    \begin{tabular}{ | m{2cm} | m{19mm} | m{19mm} | m{19mm} | m{19mm} |} 
      \hline
      Model & Lemmatization Accuracy & Levenshtein Distance & Morphological Accuracy & Morphological F-1 Score \\ 
      \hline
      SIGMORPHON & \textbf{89.03} & 0.28 & 87.63 & 94.96 \\
      \hline
      \textbf{Separate Model} & 88.84 & \textbf{0.12} & \textbf{89.66} & \textbf{96.41}  \\
      \hline
      \textbf{Sequenced Model} & 88.25 & 0.13 & \textbf{89.66} & \textbf{96.41}  \\
      \hline
    \end{tabular}
    \label{tab:table_2}
\end{table}


The terms SIGMORPHON, separate model, and sequenced model are used when Tables 1 and 2 are examined. Here, the SIGMORPHON model stands as the top achievement in the competition. The term "sequenced model" refers to the model shown in Figure 1, where the results of the morphological tagging model are input into the lemmatization model. This is not the case for the separate model, where the morphological tagging model and the lemmatization model are completely separate models. In the morphological accuracy and morphological F-1 score categories, the split model and the merged model achieved success. This is because the results of the morphological tagging model are used in both models. 

When the results are examined, the use of morphological tags as input in the lemmatization model yields different results on both datasets. As shown in Table 1, while this method increases the success in the IMST dataset, it decreases the success in the PUD dataset in Table 2. This may be because the number of records in the IMST dataset is much larger than that in the PUD dataset. If the achievements in the data sets are examined separately, the sequenced model described in Table 1 shows the highest success in all categories. When Table 2 is examined, the separate model achieves the highest success in all other categories except the lemmatization accuracy category. Looking at the results in general, the two models we presented in this study outperform the results of the SIGMORPHON competition in two datasets and in almost all categories.

\section{Conclusion}\label{sec5}

This work offers Turkish-specific morphological tagging and lemmatization models that are mindful of word meaning. Two models, the separate model and the sequenced model, depend on whether the output of the morphological tagging model is used as input or not and are presented in the lemmatization topic. The analysis of the data revealed that the morphological tags increased the lemmatization in the IMST data set but had the opposite effect in the PUD data set. The work is significant because it presents the results of the first lemmatization model specifically focused on the Turkish language. In two data sets and almost all comparison measures, the results show that this study surpasses the findings of the SIGMORPHON competition, which are the results of a multilingual study with the highest success rate shared in Turkish. The studies in the SIGMORPHON competition focused on multiple languages, whereas our study considered only one language.

\section*{Declarations}

\subsection*{Data Availability}\label{sec6}
The datasets used in this study is available at \href{https://github.com/sigmorphon/2019/tree/master/task2}{this} repository.

\subsection*{Conflict of Interest}\label{sec7}
We have no conflicts of interest to disclose.

\subsection*{Funding}
The authors received no financial support for research, authorship, and / or publication of this article.

\subsection*{Ethical approval}
This article does not contain studies with human participants or animals performed by any of the authors.

\bibliography{lemma_and_morph}


\end{document}